\documentclass[10pt,twocolumn,letterpaper]{article}

\usepackage[pagenumbers]{cvpr} % To force page numbers, e.g. for an arXiv version

\usepackage{graphicx}
\usepackage{amsmath}
\usepackage{amssymb}
\usepackage{booktabs}

\usepackage{algorithm}
\usepackage{algpseudocode}
\usepackage{hhline}
\usepackage{tabu}
\usepackage{adjustbox}
\usepackage{tabulary,multirow,overpic,xcolor}
\usepackage{color, colortbl}
\definecolor{LightGray}{gray}{0.95}

\usepackage[pagebackref,breaklinks,colorlinks]{hyperref}

\usepackage[capitalize]{cleveref}
\crefname{section}{Sec.}{Secs.}
\Crefname{section}{Section}{Sections}
\Crefname{table}{Table}{Tables}
\crefname{table}{Tab.}{Tabs.}

\def\cvprPaperID{*****} % *** Enter the CVPR Paper ID here
\def\confName{CVPR}
\def\confYear{2022}

\begin{document}

%%%%%%%%% TITLE - PLEASE UPDATE
\title{RTIC: Residual Learning for Text and Image Composition using Graph Convolutional Network}

\author{Minchul Shin~\thanks{This work is done while at NAVER/LINE Vision.} \\
Kakao Brain \\
{\tt\small min.stellastra@gmail.com}
% For a paper whose authors are all at the same institution,
% omit the following lines up until the closing ``}''.
% Additional authors and addresses can be added with ``\and'',
% just like the second author.
% To save space, use either the email address or home page, not both
\and
Yoonjae Cho, \space\space Byungsoo Ko, \space\space Geonmo Gu \\
NAVER/LINE Vision\\
{\tt\small \{yoonjae.cho92, kobiso62, korgm403\}@gmail.com}
}

\maketitle

%%%%%%%%% ABSTRACT
\begin{abstract}
In this paper, we study the compositional learning of images and texts for image retrieval. The query is given in the form of an image and text that describes the desired modifications to the image; the goal is to retrieve the target image that satisfies the given modifications and resembles the query by composing information in both the text and image modalities. To remedy this, we propose a novel architecture designed for the image-text composition task and show that the proposed structure can effectively encode the differences between the source and target images conditioned on the text. Furthermore, we introduce a new joint training technique based on the graph convolutional network that is generally applicable for any existing composition methods in a plug-and-play manner. We found that the proposed technique consistently improves performance and achieves state-of-the-art scores on various benchmarks. To avoid misleading experimental results caused by trivial training hyper-parameters, we reproduce all individual baselines and train models with a unified training environment. We expect this approach to suppress undesirable effects from irrelevant components and emphasize the image-text composition module's ability. Also, we achieve the state-of-the-art score without restricting the training environment, which implies the superiority of our method considering the gains from hyper-parameter tuning. The code, including all the baseline methods, are released~\footnote{https://github.com/nashory/rtic-gcn-pytorch}.
\end{abstract}

%%%%%%%%% INTRODUCTION
\section{Introduction}
Combining visual and contextual information, in which the desired items are retrieved from an image and a modification text that describes the desired modifications enables search engines to provide users with a powerful and intuitive experience for visual search.
Among the various approaches ~\cite{shin2019semi,zhao2017memory,ak2018learning} that have been employed to solve this problem, \textit{image-text composition} is considered as the most direct and intuitive approach.
The main principle behind this approach is to learn a non-linear mapping from a pair comprising the source image and the modification text to the target image. It indicates that the triplet of source image, target image, and text is required as the ground-truth for supervision.
The task of \textit{image-text composition}~\cite{anwaar2020compositional,vo2019composing,shin2020fashion,chen2020image,chen2020learning,jandial2020trace} has a special constraints that the output representation is required to be embedded in the image feature space and not in an arbitrary space. 
Therefore, such property should be considered carefully in designing a composition module.

\begin{figure}[t]
    \begin{center}
	    \includegraphics[width=1.0\linewidth]{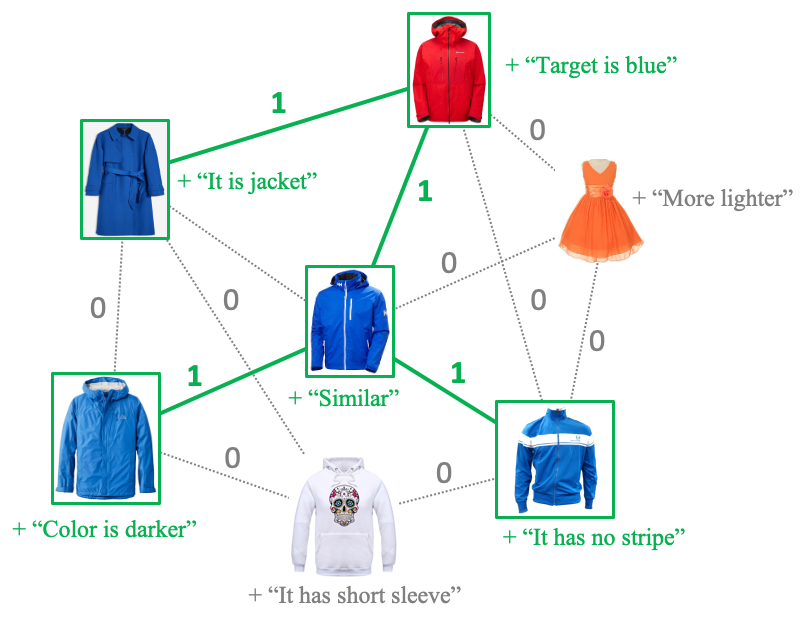}
	\end{center}
	\caption{Our concept for encoding relationships between the nodes which are defined as \textit{image-text} pairs in the graph. we consider a group of green-bordered pairs is highly correlated with one another because the pairs in the group describe visually similar target images.}
	\label{fig:teaser}
\end{figure}

In light of this, we mainly focus on the use of the skip connection~\cite{he2016deep} inspired by its powerfulness in learning representations. Despite its simplicity, little attention has been drawn to this idea in the \textit{image-text composition} task. Since the skip connection helps to learn the residual, it is intuitive to use it for encoding the differences between the source image and target image in latent space. Based on this idea, we propose a novel architecture specialized for encoding the residual between the source and target image. Our method can be considered as an advanced variant of TIRG~\cite{vo2019composing} with the integration of the skip connection.

One challenge in \textit{image-text composition} is data scarcity because collecting the ground-truth triplet pairs is extremely expensive. Therefore, it is inevitable that only a small amount of data is available for training, which causes the model to generalize poorly. To remedy this, we suggest a graph convolutional network (GCN)-based regularization technique that can be interpreted as a particular form of semi-supervised learning. We explore the possibility of utilizing graph information that encodes the similarity between \textit{image-text} pairs as shown in Figure~\ref{fig:teaser}. The proposed graph-based technique, namely GCN-stream, can be attached to any existing composition method. We found that joint training with the GCN-stream consistently improves the results of all existing methods in a plug-and-play manner.

Our ultimate goal is (1) to achieve the best score on the benchmark and (2) to compare the composition method itself as objectively as possible. The latter objective can be achieved by removing any unintended effects from the other components in the training pipeline but the composition method. We found the recent studies~\cite{anwaar2020compositional,chen2020image,chen2020learning,jandial2020trace} explore their best-performing training environment by tackling various parts in the training pipeline (\textit{e.g., word embedding, image/text encoder, composer and loss}). Although such variants may improve the result significantly, the downside is that comparing the composition methods across the papers becomes less meaningful. Therefore, we consider both aspects in our experiments: \textbf{\textit{(1) training with unified environment}} to ensure fairness and objectiveness in comparison, and \textbf{\textit{(2) training with optimal environment}} to compare the best performance of each method. We experimentally show that our proposed method can achieve state-of-the-art performance in both aspects.

\textbf{Main contributions.} In summary, the main contributions of our work are threefold: Firstly, we introduce a simple but powerful \textit{image-text composition} architecture, RTIC, that effectively uses skip connections for error encoding. Secondly, we propose a novel training technique that uses GCN as a good regularizer. The proposed technique is applicable for any existing methods in a plug-and-play manner. Lastly, we not only achieve the competitive result on various benchmarks but also provide a thorough and objective comparison between the composition methods by performing experiments with the \textit{unified} and \textit{optimal} environment both.

%%%%%%%%% RELATED WORK
\section{Related Work}
\noindent
\textbf{Image and Text Composition.}
Many studies~\cite{ben2019block,ben2017mutan,kim2016hadamard,yu2017multi,yu2018beyond,fukui2016multimodal} have investigated effective ways to combine more than two different modalities in \textit{feature fusion}. The \textit{image-text composition} task can be seen as a unique form of feature fusion in that the textual information describes the target image to be retrieved. Inspired by the release of public datasets such as Fashion-IQ~\cite{guo2019fashion}, many studies have explored this problem. In their pioneering work, Vo \textit{et al.}~\cite{vo2019composing} proposed TIRG, which uses a gating mechanism to determine the channels of the image representation to be modified conditioned by a text. Based on their work, various methods~\cite{anwaar2020compositional,vo2019composing,shin2020fashion,chen2020image,chen2020learning,jandial2020trace} that achieved state-of-the-art results on this task have been reported. VAL~\cite{chen2020image} employs a hierarchical feature matching that exploits the intermediate features from the image encoder. TRACE~\cite{jandial2020trace} also uses a complicated hierarchical feature aggregation technique and applies three different loss functions simultaneously. ComposeAE\cite{vo2019composing} suggests a novel embedding space that can semantically tie representations from the text and image modalities.
The winning solutions~\cite{shin2020fashion,kim2021cycled} for the competitions employed more exhaustive tricks to achieve the best performance in this task by using careful hyperparameter tuning and model ensemble with Bayes optimization~\cite{snoek2012practical} to improve the results significantly.
Although the methods mentioned above have focused on achieving the highest score on the benchmark, only a few studies have focused on the composition module itself~\cite{vo2019composing,anwaar2020compositional}. Since different aspects (\textit{e.g., image/text encoder or loss function}) of image and text composition were addressed in recent related works~\cite{anwaar2020compositional,shin2020fashion,chen2020image,chen2020learning,jandial2020trace}, a naive comparison could lead to a misinterpretation. Therefore, we aim to achieve the highest score on the benchmark and, at the same time, analyze the ability of the composition module as objectively as possible.

\begin{figure*}[t!]
    \begin{center}
	    \includegraphics[width=1.0\linewidth]{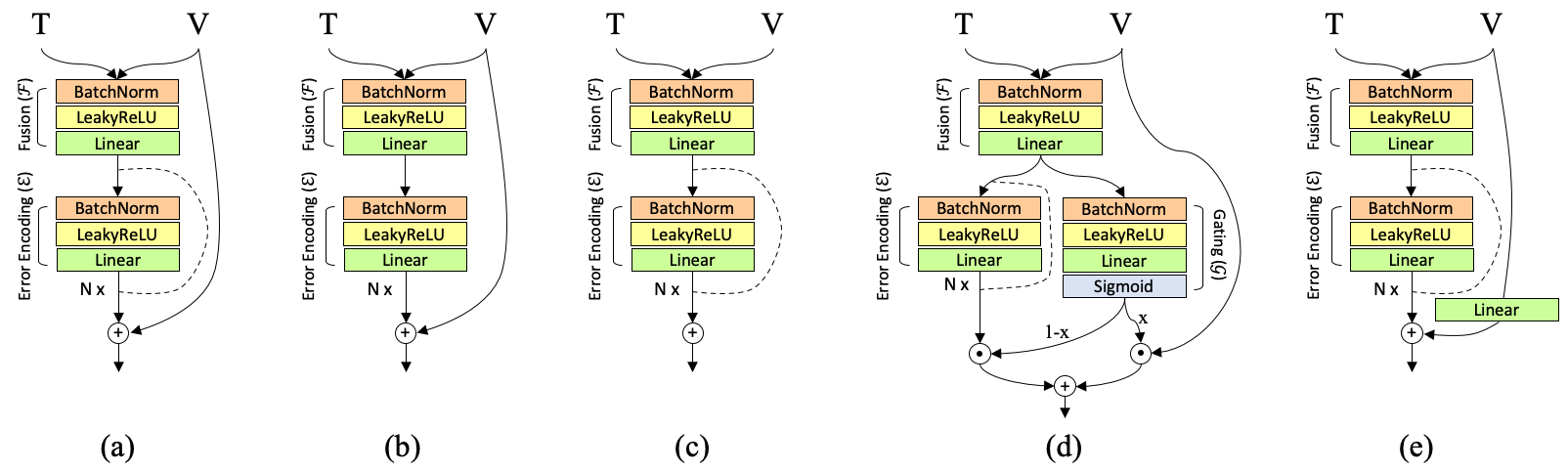}
	\end{center}
	\caption{Alternative models are examined to justify our architecture design for the composition method. $T$ and $V$ indicate the representations extracted from the text and the image encoder, respectively. We conceptually illustrate the composition of layers for each blocks. The detailed composition is described in the appendix.}
	\label{fig:variants}
\end{figure*}

\noindent
\textbf{Graph Convolutional Network.}
The graph convolutional network (GCN) was first introduced in \cite{kipf2016semi} to perform semi-supervised classification. The essential strength of the GCN is its ability to extract meaningful information encoded in the form of a structured graph from the relationships between the nodes. While most applications of GCN were limited to node classification in the early stage, many efforts~\cite{chen2019multi,liu2020od,chen2020hierarchical,liu2019guided} have been made to use appropriately modified GCN as a joint training module. In the most closely related work to ours, Chen \textit{et al.}~\cite{chen2019multi} applied GCN for multi-label classification by learning linear classifiers from the graph for projecting features into class labels. In that work, the word embeddings were used for the node feature matrix, and the label co-occurrence was measured to determine the values for the graph edges. Although the effectiveness of GCN has been proved in many other studies~\cite{you2020l2,wei2020view,cheng2020skeleton}, applying GCN for a new task such as \textit{image-text composition} is not trivial. The main reason is that the GCN requires a high-quality graph. Building such meaningful graphs is tricky because a proper definition of the graph nodes and a careful strategy for encoding the relationships between the nodes are needed. In this paper, we introduce the application of a GCN for the \textit{image-text composition} task. We define the nodes as \textit{image-text} pairs and assign higher correlations to nodes that have target images with high visual similarity. We explain our proposed approach in detail in the following section.

%%%%%%%%% PROPOSED APPROACH
\section{Proposed Approach}
This section introduces our proposed method called the Residual Text and Image Composer (RTIC). The goal of the \textit{image text composition} task is to combine the source image $I^{src}$ and the text $T$ by using a multi-modal composer $\psi(\cdot, \cdot)$ to generate a desired representation close to the representation of the target image $I^{trg}$. For this purpose, $\{I^{src}$, $T$, and $I^{trg}\}$ triplets are required for supervision. Given a composition method such as RTIC, we denote the conventional training pipeline in which pairwise ranking loss (\textit{e.g., triplet loss}) is applied on the ground-truth triplets as the main-stream. The proposed GCN-stream is an auxiliary module attached to the main-stream. Our approach is based on two-stage training. In the first stage, the multi-modal composer, RTIC, is trained in the main-stream. Then, in the second stage, a graph is built using the trained RTIC to configure the GCN-stream and train the main-stream model jointly with the GCN-stream. The weights in the main-stream are initialized from a pre-trained composer or a scratch model for the composer. The advantage of training from scratch is that any type of composer different from the first stage can be chosen because the pre-trained composer is no longer required once the graph is constructed. On the other hand, more performance gain can be achieved when the weights of the pre-trained composer are transferred, while the composer architecture is restricted to be the same used at the first stage.

\begin{figure*}[t]
    \begin{center}
	    \includegraphics[width=1.0\linewidth]{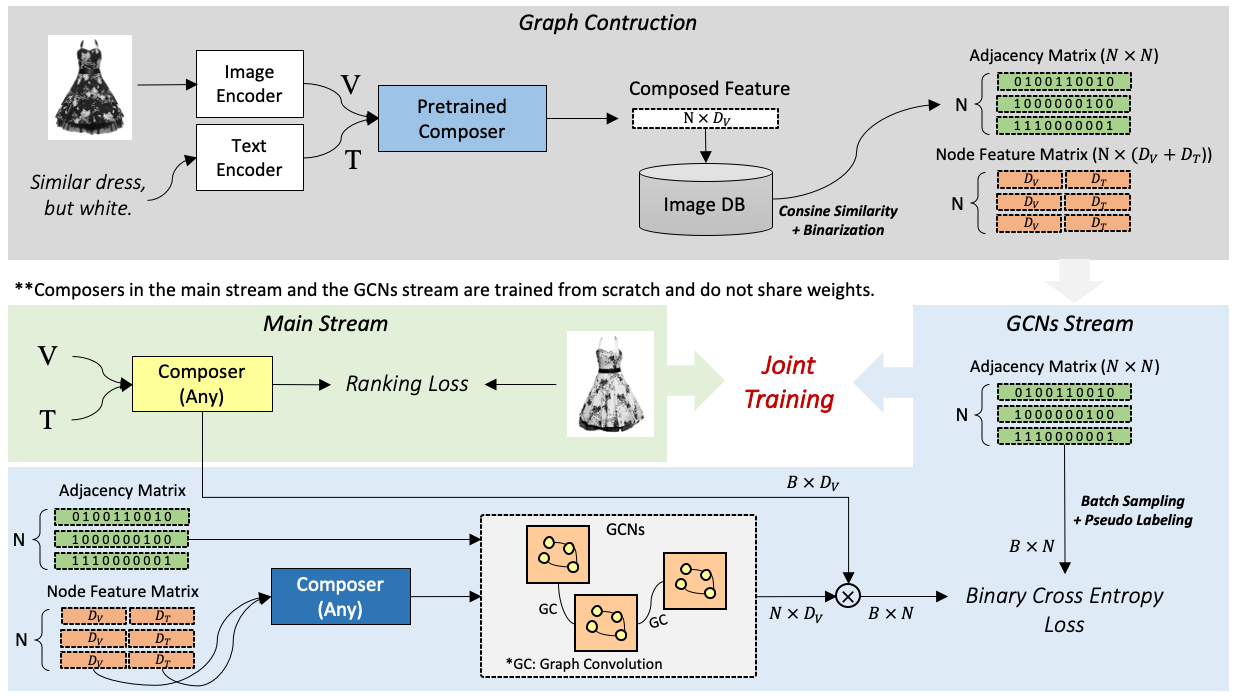}
	\end{center}
	\caption{The pipeline for joint training with the GCN-stream. The blue region indicates the GCN module, which can be attached to any existing composition method in a plug-and-play manner. Note that the weights between the two composers in each stream are not shared. We use notations $\hat{A}$ for the GCN input and $A'$ for the batch sampling and pseudo-labeling.}
	\label{fig:main_diagram}
\end{figure*}

\subsection{Residual Text and Image Composer}
RTIC is designed to learn the residual between the target and source image representations. The idea is that, given the images $I^{src}$ and $I^{trg}$ and the features $v^{src}$ and $v^{trg}$ of the source and the target, respectively, we consider the target $v^{trg}$ as an addition of the source $v^{src}$ and the residual $h$ conditioned on the text feature $t$, which is represented as $h = \delta(v^{src}; t)$, where $\delta$ is a nonlinear mapping. Therefore, the final composed feature $\Tilde{v}$ is formulated as $\Tilde{v} = v^{src} + h \simeq \psi(v^{src}; t) = v^{trg}$, assuming that $\psi(\cdot, \cdot)$ is an ideal composition module. In the RTIC, the residual $h$ is considered as an error between the two different image features, $v^{src}$ and $v^{trg}$, which is continuously distributed over the entire channel. The RTIC consists of three main components: a fusion block, an error encoding block, and a gating block. The fusion block $\mathcal{F}$ combines two modalities, $t$ and $v^{src}$, into a single representation $x = \mathcal{F}(v^{src}, t)$. Then, $x$ is fed to the error encoding block $\mathcal{E}$ and gating block $\mathcal{G}$ to obtain the final composed feature $\tilde{v}$ as follows:
\begin{equation}
    \label{eq:rtic_eq}
    \begin{aligned}
        & s = \mathcal{G}(\mathcal{F}(v^{src}, t)) = \mathcal{G}(x), \\
        & \mathcal{X}^n = \mathcal{X}^{n-1} + \mathcal{E}^{n}(\mathcal{X}^{n-1}), \; \; \mathcal{X}^{0} = x, \\
        & \tilde{v} = (1-s) \cdot \mathcal{X}^n + s \cdot v^{src} \simeq v^{trg} , \\
    \end{aligned}
\end{equation}
where $\mathcal{X}^n$ represents the output of the $n$-$th$ error encoding block $\mathcal{E}^{n}$. Note that the output of the previous error encoding block $\mathcal{X}^{n-1}$ is added to the next output using the skip connection. Figure~\ref{fig:variants} conceptually illustrates the flow to facilitate understanding. In all our experiments, $n$ was set to four by default. For the next step, we follow the general pipeline of the other existing methods for the main-stream, in which a pairwise ranking loss is applied using $\tilde{v}$ as the anchor and $v^{trg}$ as the positive. Following~\cite{vo2019composing}, we use the DML loss ($K$=$B$) for the pairwise ranking loss. Although we observed that a better overall score was achieved using the batch hard triplet loss~\cite{hermans2017defense}, we intentionally used the DML loss because such a classification-based objective can effectively suppress undesirable effects caused by the sampling strategy. We refer interested readers to ~\cite{vo2019composing} for detailed definition of DML Loss. The structure finally chosen for the RTIC is architecture \textit{(d)} in Figure~\ref{fig:variants}. The details of our design choice are discussed in the later section.

In summary, RTIC has two main differences from TIRG. First, RTIC uses channel-wise linear interpolation between the source image feature $v^{src}$ and the residual $h$ in which the sigmoid function is applied to the output of the gating block $\mathcal{G}$ to determine the ratio. Second, RTIC employs a deeper architecture for error encoding by stacking $n$ learning blocks in series using skip connections. We demonstrate the potential of our idea by winning second place in the competition~\footnote{https://sites.google.com/view/cvcreative2020/fashion-iq}. The following section explains our proposed graph-based approach, RTIC-GCN, a further improved version of RTIC.

\subsection{GCN as Good Regularizer}

One challenge in the \textit{image-text composition} is that an infinite number of perceptually acceptable image and text pairs exist, while the ground-truth triplets ($I^{src}$, $I^{trg}$, and $T$) in the training set reflect only a tiny fraction of these pairs. Such a scarcity of training pairs causes poor generalization. To resolve this, we introduce a novel method of using a graph convolutional neural network (GCN) as a regularizer by propagating information between adjacent neighbors. Our intuition is that, because there exists an unlimited number of ways to describe $I^{trg}$ given $I^{src}$, we consider $\{I^{src}, T\}$ pairs as a group that describes a visually similar target image $I^{trg}$. The $\{I^{src}, T\}$ pairs in the same group are hence highly correlated with one another. For example,  an image of a \textit{red coat} with the text ``it is blue’’ and an image of a \textit{blue shirt} with the text ``it is a coat’’ would both have an image of a \textit{blue coat} as the target, which means that the two pairs are highly correlated. We model such relationships between pairs in the form of a graph by measuring the visual similarities between the target images of the triplets. We conceptually illustrate our intuition in Figure~\ref{fig:teaser}.

\subsubsection{Revisiting GCN}
Before moving into the details, we briefly review how a GCN works. The core idea in a GCN is that the node representation is updated by propagating information from the neighboring nodes based on graph-structured data. The goal of the GCN is to learn a nonlinear mapping $f(\cdot, \cdot)$ on the graph. The $l$-$th$ layer of the GCN takes the node feature matrix $H^{l} \in R^{N \times D}$ and the correlation matrix $\Lambda \in R^{N \times N}$ as inputs, where $N$ and $D$ indicate the number of nodes and the dimensionality of the node features, respectively. The operation on each layer of the GCN can be represented as
\begin{equation}
    \label{eq:gcn_revisited}
    \begin{aligned}
        & \hat{\Lambda} = D^{\frac{1}{2}}\Lambda D^{-\frac{1}{2}}, \\
        & H^{l+1} = h(\hat{\Lambda} H^{l} W^{l}), \\
    \end{aligned}
\end{equation}
where $D$ represents a diagonal matrix of $\Lambda$ and $W^{l}$ indicates the transformation matrix of the $l$-$th$ layer to be learned. We used ReLU for the activation function $h$. For more details, we refer interested readers to ~\cite{kipf2016semi}.

\subsubsection{Graph construction}
Most importantly, we define each graph node as an \textit{image-text} pair, which means that the $N$ ground-truth pairs in the training data create $N$ graph nodes in total. Assuming that we have a learned image encoder $\phi_{pretrained}$ trained in the first stage, we model the correlations between the nodes (\textit{image-text} pairs) by measuring the cosine similarities between the target features of the corresponding nodes $\mathcal{V} = \{v^{trg}_{1}, ... , v^{trg}_{N}\}$, where the target features of the $i$-$th$ node are obtained by $v^{trg}_{i} = \phi_{pretrained}(I_{i}^{trg})$. As a result, we obtain a symmetrical correlation matrix $\Lambda \in R^{N \times N}$ because the cosine similarity measure has a commutative property, and the similarity with the self-node is always 1. Then, we binarize $\Lambda$ with the threshold $\tau$ to filter out the noisy edges:
\begin{equation}
    \label{eq:binarized}
    \Lambda'_{ij} =
    \begin{cases}
        1, & \text{if } \Lambda_{i,j} \geq \tau \\
        0, & \text{if } \Lambda_{i,j} < \tau, \\
    \end{cases}
\end{equation}
where $\tau$ is a hyperparameter set to the value that makes the average number of activated edges in each node 15\% of $N$. Chen \textit{et al.}~\cite{chen2019multi} addressed the over-smoothing problem that occurs when each node feature is repeatedly updated using its own features and those of the neighboring nodes by employing a re-weighting strategy. We follow the same scheme with a slight modification:
\begin{equation}
    \label{eq:reweighted}
    {\Lambda''}_{ij} =
    \begin{cases}
        1, & \text{if } i = j\\
        0.33 / \Sigma_{j=1, i \neq j}^{N}\Lambda'_{ij}, & \text{if } i \neq j. \\
    \end{cases}
\end{equation}

Finally, $ \Lambda''$ is re-normalized: $\hat{\Lambda} = D^{\frac{1}{2}}\Lambda'' D^{- \\ \frac{1}{2}}$, as described in Equation~ \ref{eq:gcn_revisited}.
The full graph construction algorithm is given in Algorithm~\ref{algo:graph_construction}.

\definecolor{light-gray}{gray}{0.5}  
\begin{algorithm}
	\caption{Graph construction algorithm for encoding the contextual similarity between the nodes represented by the \textit{image}-\textit{text} pairs} 
	\begin{algorithmic}[1]
	    \State \textbf{Input}: Image encoder $\phi$, text encoder $\lambda$, $N$ triplets of image and text $X = \{(I^{src}_i, T_i, I^{trg}_i) : i \in (1,...,N) \}$.
	    \State $(\mathcal{H}, \Omega) \leftarrow$ ([], [])
	    \For {$i = 1$ \textbf{to} $N$ }
	        \State \textcolor{light-gray}{// extract image and text representations}
	        \State $v^{src}_i \leftarrow \phi(I^{src}_i)$, $t_i \leftarrow \lambda(T)$, $v^{trg}_i \leftarrow \phi(I^{trg}_i)$
	        \State $\mathcal{H}.append(v^{src}_i \oplus t_i)$ \textcolor{light-gray}{// node feature matrix}      
	        \State $\Omega.append(v^{trg})$ \textcolor{light-gray}{// create feature DB}
	    \EndFor
	    \State $\Lambda \leftarrow cosine\_similarity(\Omega \cdot \Omega^T)$ \textcolor{light-gray}{// correlation matrix}
	    \State $\Lambda' \leftarrow binarized(\Lambda)$ \textcolor{light-gray}{// thresholding}
	    \State $\Lambda'' \leftarrow reweighted(\Lambda')$ \textcolor{light-gray}{// to avoid over-smoothing}
	    \State $\hat{\Lambda} \leftarrow normalized(\Lambda'')$ \textcolor{light-gray}{// to keep the scale}
	    \State \textbf{Output}: Normalized correlation matrix $\hat{\Lambda} \in R^{N \times N}$, node feature matrix $\mathcal{H} \in R^{N \times (D_V + D_T)}$.
	\end{algorithmic}
	\label{algo:graph_construction}
\end{algorithm}

    %% Table 1: Benchmark score on fashion-iq
    \begin{table*}[t!]
    \caption{The performance evaluated on the Fashion-IQ. The scores in white-colored rows indicate the comparison between the different composition methods when trained and evaluated with our constrained environment. The scores with the optimal environment are brought from each method's paper to compare the best performance. $\textsuperscript{\dag}$ mark indicates that the weights of the pre-trained composer are transferred.}
    \centering
    \begin{adjustbox}{width=1.0\textwidth}
    \begin{tabular}{cccccccc}
        \toprule
        \multirow{2}{*}{Method} & Average & \multicolumn{2}{c}{Shirt} & \multicolumn{2}{c}{Dress} & \multicolumn{2}{c}{Toptee} \\ \cline{3-8}
        & (R@10 + R@50) / 2 & R@10 & R@50 & R@10 & R@50 & R@10 & R@50 \\
        \hline \hline
        \rowcolor[gray]{0.85}\multicolumn{8}{l}{\textit{\textbf{Training with unified environment (objective comparison)}}} \\
        \hline
        Param Hashing~\cite{noh2016image} & $26.54 \pm 0.20$ & $12.26 \pm 0.58$ & $32.83 \pm 0.66$ & $15.48 \pm 0.53$ & $39.13 \pm 0.52$ & $17.66 \pm 0.39$ & $41.89 \pm 0.68$ \\
        MRN~\cite{kim2016multimodal} & $28.83 \pm 0.18$ & $14.50 \pm 0.43$ & $35.43 \pm 0.39$ & $16.60 \pm 0.47$ & $40.50 \pm 0.76$ & $20.22 \pm 0.55$ & $45.74 \pm 0.26$ \\
        FiLM~\cite{perez2018film} & $28.72 \pm 0.32$ & $13.56 \pm 0.33$ & $35.64 \pm 0.94$ & $16.78 \pm 1.04$ & $40.77 \pm 0.98$ & $19.91 \pm 0.62$ & $45.64 \pm 0.60$ \\
        TIRG~\cite{vo2019composing} & $30.71 \pm 0.25$ & $16.12 \pm 0.39$ & $37.69 \pm 0.70$ & $19.15 \pm 0.61$ & $43.01 \pm 0.91$ & $21.21 \pm 0.70$ & $47.08 \pm 0.49$ \\
        ComposeAE~\cite{anwaar2020compositional} & $19.61 \pm 0.41$ & $9.96 \pm 0.60$ & $25.14 \pm 0.88$ & $10.77 \pm 0.70$ & $28.29 \pm 0.49$ & $12.74 \pm 0.67$ & $30.79 \pm 0.62$ \\
        VAL~\cite{chen2020image} & $27.16 \pm 0.24$ & $13.62 \pm 0.59$ & $33.81 \pm 0.95$ & $16.03 \pm 0.58$ & $39.07 \pm 0.64$ & $18.02 \pm 0.52$ & $42.40 \pm 0.44$ \\
        \hline
        Block~\cite{ben2019block} & $27.84 \pm 0.34$ & $13.67 \pm 0.69$ & $35.35 \pm 0.77$ & $16.01 \pm 0.54$ & $39.61 \pm 0.56$ & $18.39 \pm 0.74$ & $44.03 \pm 0.58$ \\
        Mutan~\cite{ben2017mutan} & $29.34 \pm 0.22$ & $15.20 \pm 0.82$ & $36.17 \pm 0.42$ & $17.46 \pm 0.56$ & $42.14 \pm 0.46$ & $20.04 \pm 0.62$ & $45.03 \pm 0.62$ \\
        MLB~\cite{kim2016hadamard} & $27.12 \pm 0.49$ & $13.30 \pm 0.44$ & $33.16 \pm 0.43$ & $15.33 \pm 0.50$ & $39.71 \pm 0.82$ & $17.75 \pm 0.40$ & $43.51 \pm 1.10$ \\
        MFB~\cite{yu2017multi} & $27.98 \pm 0.21$ & $13.94 \pm 0.67$ & $34.37 \pm 0.38$ & $16.60 \pm 0.58$ & $40.36 \pm 0.90$ & $18.36 \pm 0.56$ & $44.24 \pm 0.65$ \\
        MFH~\cite{yu2018beyond} & $27.96 \pm 0.27$ & $13.87 \pm 0.59$ & $35.04 \pm 0.22$ & $15.95 \pm 0.61$ & $40.16 \pm 0.84$ & $18.52 \pm 0.63$ & $44.21 \pm 0.75$ \\
        MCB~\cite{fukui2016multimodal} & $29.28 \pm 0.29$ & $14.77 \pm 0.75$ & $36.34 \pm 0.48$ & $17.43 \pm 0.57$ & $41.69 \pm 0.63$ & $19.96 \pm 0.75$ & $45.46 \pm 1.01$ \\
        \hline
        RTIC & $31.28 \pm 0.22$ & $16.93 \pm 0.45$ & $38.36 \pm 0.61$ & $19.40 \pm 0.50$ & $43.51 \pm 0.90$ & $21.58 \pm 0.70$ & $47.88 \pm 0.90$ \\
        RTIC-GCN & \textbf{31.68} $\pm$ \textbf{0.30} & \textbf{16.95} $\pm$ \textbf{0.59} & \textbf{38.67} $\pm$ \textbf{0.74} & \textbf{19.79} $\pm$ \textbf{0.41} & \textbf{43.55} $\pm$ \textbf{0.24} & \textbf{21.97} $\pm$ \textbf{0.71} & \textbf{49.11} $\pm$ \textbf{0.87} \\
        \hline
        \rowcolor[gray]{0.85}\multicolumn{8}{l}{\textit{\textbf{Training with optimal environment (best performance)}}} \\
        \hline
        JVSM~\cite{chen2020learning} & $ 19.26 $ & $ 12.00 $ & $ 27.10 $ & $ 10.70 $ & $ 25.90 $ & $ 13.00 $ & $ 26.90 $ \\
        TRACE w/ BERT~\cite{jandial2020trace} & $ 34.38 $ & $ 20.80 $ & $ 40.80 $ & $ 22.70 $ & $ 44.91 $ & $ 24.22 $ & $ 49.80 $ \\
        VAL w/ GloVe~\cite{chen2020image} & $ 35.38 $ & $ 22.38 $ & $ 44.15 $ & $ 22.53 $ & $ 44.00 $ & $ 27.53 $ & $ 51.68 $ \\
        CIRPLANT w/ OSCAR~\cite{cirplant} & $ 30.20 $ & $ 17.53 $ & $ 38.81 $ & $ 17.45 $ & $ 40.41 $ & $ 21.64 $ & $ 45.38 $ \\
        MAAF~\cite{MAAF} & $ 36.60 $ & $ 21.30 $ & $ 44.20 $ & $ 23.80 $ & $ 48.60 $ & $ 27.90 $ & $ 53.60 $ \\
        CurlingNet~\cite{yu2020curlingnet} & $ 38.45 $ & $ 21.45 $ & $ 44.56 $ & $ 26.15 $ & $ \underline{53.24} $ & $ \underline{30.12} $ & $ 55.23 $ \\
        CoSMo~\cite{lee2021cosmo} & $ 39.45$ & $ \textbf{24.90} $ & $ \textbf{49.18} $ & $ 25.64 $ & $ 50.30 $ & $ 29.21 $ & $ 57.46 $ \\
        \hline
        RTIC w/ GloVe & $ 39.22 $ & $ 22.81 $ & $ 45.97 $ & $ 27.81 $ & $ 53.15 $ & $ 30.04 $ & $ 55.53 $ \\
        RTIC-GCN w/ GloVe & $ \underline{39.55} $ & $ 23.26 $ & $ 45.68 $ & $ \underline{27.96} $ & $ 52.70 $ & $ 29.98 $ & $ \underline{57.73} $ \\
        RTIC-GCN w/ GloVe$^\textsuperscript{\dag}$ & $ \textbf{40.64} $ & $ \underline{23.79} $ & $ \underline{47.25} $ & $ \textbf{29.15} $ & $ \textbf{54.04} $ & $ \textbf{31.61} $ & $ \textbf{57.98} $ \\
        \bottomrule
    \end{tabular}
    \end{adjustbox}
    \label{tab:bechmark_fashion_iq}
    \end{table*}

\noindent
\textbf{Joint training with GCN-stream.}
Denoting the general training pipeline (\textit{e.g., training RTIC with ranking loss}) as the main-stream, the GCN-stream can be defined as an auxiliary module that aims to improve the mainstream model by injecting additional signals in a semi-supervised manner. Once the graph is constructed, the GCN-stream can be configured and jointly trained with the main-stream. The composers in each stream in Figure~\ref{fig:main_diagram} are randomly initialized, and the graph is the only source of information that the model can benefit from during training. The GCN-stream is configured using the output correlation matrix $\hat{\Lambda} \in R^{N \times N}$ and the node feature matrix $\mathcal{H} \in R^{N \times (D_V + D_T)}$, where $N$ is the number of $\{I^{src}, T\}$ pairs, and $D_V$ and $D_T$ are the output feature dimensions of the image and the text encoders, respectively. For the next step, the image and text feature matrices $\Bar{v} \in R^{N \times D_V}, \Bar{t} \in R^{N \times D_T}$ are split from $\mathcal{H}$ and fed to a randomly initialized new composer in the GCN-stream that converts $\mathcal{H}$ into a matrix of composed features $\tilde{\mathcal{H}} \in R^{N \times D_V}$. Note that the weights of the composers in each stream are not shared because the two composers will have different input distributions as the training proceeds. Finally,  $\tilde{\mathcal{H}}$ and $\hat{\Lambda}$ are forwarded by the GCN layers to obtain the GCN-stream output $\mathcal{W} \in R^{N \times D_V}$.

We inject information from the GCN-stream into the main-stream in the form of node classifications~\cite{chen2019multi}. Given a mini-batch $B$, we perform matrix multiplication between the output of the main-stream $\mathcal{X}=\{\tilde{v_1}, .. , \tilde{v_B}\} \in R^{B \times D_V}$ and the output of the GCN-stream $\mathcal{W} \in R^{N \times D_V}$ to obtain the logits $x = \mathcal{X} \times \mathcal{W}^T \in R^{B \times N}$. Then, we use the binary cross-entropy loss to perform a node classification on the $N$ graph nodes. This process can be seen as finding the most relevant neighbors for a batch of queries among all the $\{I^{src}, T\}$ pairs that exist in the graph. For supervision, we sample $B$ as the number of one-hot vectors $c \in R^{B \times N}$ on the fly from $\Lambda'$ according to the unique key assigned to the pairs in the mini-batch. The terminologies---batch sampling and pseudo labeling---in this process are represented in Figure~\ref{fig:main_diagram}. Finally, the loss for the joint training is obtained as the sum of the pairwise ranking loss in the main-stream and the binary classification loss in the GCN-stream, $\mathcal{L}_{final} = \lambda_{pair} \cdot \mathcal{L}_{pair} + \lambda_{bce} \cdot \mathcal{L}_{bce}$, where ($\lambda_{pair}, \lambda_{bce}$) = (1.0, 1.0) is used by default. Although the GCN-stream requires additional GPU memory, as shown in Figure~\ref{fig:gpu_memory}, the GCN-stream is used only for the training and is removed for the inference. Importantly, it means that our method requires no additional model parameters for the inference.

%%%%%%%%% EXPERIMENT
\section{Experiment}

\subsection{Experimental Settings}
In this section, We mainly discuss our training environment. We also provide more explanations about the other trivial training settings in the appendix. While such pre-processings are not our primary focus, we claim that having an identical procedure is highly important in reproducing the result.

\subsubsection{Training Environment}
We train our model on two environments both: \textit{unified} environment, \textit{optimal} environment.
For \textit{unified} environment, we unified training conditions across the different composition methods. We keep each component (\textit{e.g., image/text encoder, loss function, image transform, etc.}) in the pipeline as simple as possible to maximally suppress the misleading performance gain from parameter optimization. All the ablations in our experiments are performed in this environment. While such a constrained condition degrades the overall benchmark scores, it makes the results more reliable and objective for comparison. For example, we used the DML Loss~\cite{vo2019composing} for the ranking loss because it is independent of the sampling strategy. No initialization such as GloVe or BERT was used for the word embeddings because the effect of the initialization is highly coupled with the type of tokenizer used and the quality of the pre-trained word embeddings. We also used SGD with a fixed $lr$=0.01 for the optimizer with no parameter tuning because adjusting $lr$ can increase the overall score significantly, as described in Table~\ref{tab:ablation_study}. On the other hand, evaluating with \textit{unified} environment only can overlook the fact that different methods need their optimization strategy. To remedy this, we also trained our model with \textit{optimal environment}. The best parameters for training RTIC is found exhaustively and compared with the best scores of the other methods reported in their papers. We report the average score of eight trials for all our experiments to obtain more precise and more evident trends. The details of each training environment are described in the appendix.

\subsection{Performance Evaluation}

\noindent
\textbf{Comparison with the state-of-the-art.}
For evaluation, we measured the top-k recall (R@K) on different benchmarks with the unified environment. While following our settings led to relatively low scores, the composition methods can be compared fairly in this constrained environment. Table~\ref{tab:bechmark_fashion_iq} and Table~\ref{tab:bechmark_shoes_birds} show the results for the three benchmarks. We intentionally omitted some recent methods~\cite{chen2020learning,jandial2020trace,chen2020image,yu2020curlingnet} because these works do not solely focus on the composition method itself but also on other aspects such as the image/text encoders and loss functions. Instead, we compared our method with other methods trained with our own best settings. Although this approach might not be sufficiently objective, it would still provide some naive insights.

Table~\ref{tab:bechmark_fashion_iq} shows that RTIC already achieves the highest score among the various methods and that joint training with the GCN-stream (RTIC-GCN) further improves the score by approximately 0.4\%p. The scores for the methods marked by $\textsuperscript{\dag}$ were obtained using the best settings for the methods reported in their respective papers. To achieve the state-of-the-art score, we gradually adjusted the detailed training parameters, as described in Table~\ref{tab:ablation_study}. We did not employ trivial tricks that might improve the score, such as aggregating multiple local features or concatenating the intermediate features, because such tricks excessively increase the memory usage and model complexity. Nevertheless, we achieved a 2.19\%p improvement compared to the previous state-of-the-art.

    %% Table 2: Benchmark score on shoes and birds
    \begin{table}[h]
    \caption{Benchmark scores on Shoes and Birds-to-Words dataset.}
    \centering
    \begin{adjustbox}{width=0.48\textwidth}
    \begin{tabular}{ccccc}
        \toprule
        \multirow{2}{*}{Method} & \multicolumn{2}{c}{Shoes} & \multicolumn{2}{c}{Birds-to-Words} \\ \cline{2-5}
        & R@10 & R@50 & R@10 & R@50 \\
        \hline \hline
        Param Hashing & $ 36.37 \pm  0.47 $ & $ 67.24 \pm 0.88 $ & $ 27.54 \pm 1.57 $ & $ 58.43 \pm 1.59 $ \\
        MRN & $ 37.51 \pm 0.47 $ & $ 68.66 \pm 0.75 $ & $ 37.33 \pm 1.38 $ & $ 68.97 \pm 1.35 $ \\
        FiLM & $ 36.74 \pm 0.56 $ & $ 68.47 \pm 0.43 $ & $ 32.58 \pm 1.59 $ & $ 65.80 \pm 1.48 $ \\
        TIRG & $40.40 \pm 0.69$ & $70.65 \pm 0.65$ & $ 33.83 \pm 1.53 $ & $ 65.96 \pm 1.56 $ \\
        ComposeAE & $ 31.25 \pm 0.67 $ & $ 60.30 \pm 0.33 $ & $ 29.60 \pm 0.99 $ & $ 59.82 \pm 1.83 $ \\
        VAL & $ 36.91 \pm 0.75 $ & $ 67.16 \pm 0.73 $ & $ 28.65 \pm 1.48 $ & $ 60.22 \pm 1.31 $ \\
        \hline
        Block & $ 36.82 \pm 0.61 $ & $ 67.79 \pm 0.51 $ & $ 26.19 \pm 1.63 $ & $ 57.89 \pm 1.37 $ \\
        Mutan & $ 39.13 \pm 0.73 $ & $ 69.86 \pm 0.53 $ & $ 32.15 \pm 1.83 $ & $ 62.36 \pm 1.59 $ \\
        MLB & $ 35.35 \pm 0.88 $ & $ 67.30 \pm 0.57 $ & $ 30.93 \pm 1.77 $ & $ 63.22 \pm 1.72 $ \\
        MFB & $ 36.59 \pm 1.54 $ & $ 67.58 \pm 0.49 $ & $ 30.43 \pm 1.92 $ & $ 64.03 \pm 1.69 $ \\
        MFH & $ 35.85 \pm 1.20 $ & $ 68.07 \pm 1.07 $ & $ 31.11 \pm 2.18 $ & $ 64.18 \pm 1.93 $ \\
        MCB & $ 38.70 \pm 0.66 $ & $ 68.96 \pm 0.69 $ & $ 30.86 \pm 1.44 $ & $ 62.80 \pm 2.12 $ \\
        \hline
        RTIC & \textbf{43.66} $\pm$ \textbf{0.67} & \textbf{72.11} $\pm$ \textbf{0.51} & 37.40 $\pm$ 1.36 & 66.97 $\pm$ 1.70 \\
        RTIC-GCN & 43.38 $\pm$ 0.88 & 72.09 $\pm$ 0.48 & \textbf{37.56} $\pm$ \textbf{1.12} & \textbf{67.72} $\pm$ \textbf{1.46} \\
        \bottomrule
    \end{tabular}
    \end{adjustbox}
    \label{tab:bechmark_shoes_birds}
    \end{table}

\noindent
\textbf{Architecture verification.} In Table~\ref{tab:structure_analysis}, we verify our architecture design by comparing the possible variants of the RTIC structure and analyzing the effect of each component. The best structure is \textit{(d)}, which consistently shows the best score, followed by \textit{(a)} with a marginal gap. The main feature of architecture \textit{(d)} is the use of a channel-wise gating mechanism that performs a linear interpolation between the image feature and the residual. Interestingly, the performance drops significantly without the skip connections inside the error encoding stack (\textit{(a) vs. (b)}). The score degradation caused by removing the skip connections from the visual representation is relatively small (\textit{(a) vs. (c)}). Adjusting the latent space of the image by attaching an additional projection layer is not helpful (\textit{(a) vs. (e)}). Finally, \textit{(d)} was chosen for the final architecture design of the RTIC.

    % Table 3. Structure analysis
    \begin{table}[h]
    \caption{Comparison between alternative models (\textit{a-e}) on \textit{image-text composition} task. The compared structures are illustrated in Figure~\ref{fig:variants}.}
    \centering
    \begin{adjustbox}{width=0.40\textwidth}
    \begin{tabular}{cccc}
        \toprule
              & \multicolumn{3}{c}{(R@10 + R@50) / 2} \\ \cline{2-4} 
              & Fashion IQ   & Shoes   & Birds-to-Words   \\ \hline \hline
        (a)   & $31.28 \pm 0.22$ & $57.58 \pm 0.40$ & $51.73 \pm 0.49$ \\
        (b)   & $24.71 \pm 0.69$ & $53.07 \pm 0.51$ & $46.21 \pm 1.84$ \\
        (c)   & $30.44 \pm 0.23$ & $55.43 \pm 0.60$ & $48.38 \pm 0.40$ \\
        (d)   & \textbf{31.41 $\pm$ 0.41} & \textbf{57.73 $\pm$ 0.45} & \textbf{52.24 $\pm$ 1.06} \\
        (e)   & $30.02 \pm 0.25$ & $55.81 \pm 0.54$ & $48.42 \pm 1.55$ \\
        \bottomrule
    \end{tabular}
    \end{adjustbox}
    \label{tab:structure_analysis}
    \end{table}

\noindent
\textbf{Effect of graph quality.} We examine the effect of the graph quality on the joint training with the GCN-stream in Table~\ref{tab:graph_quality}. Assuming that the pre-trained RTIC can construct more precise graphs than the other methods based on the results in Table~\ref{tab:bechmark_fashion_iq}, we compare the improvements of four existing methods trained with the GCN-stream (\textit{TIRG-GCN, MRN-GCN, ComposeAE-GCN, Param Hashing-GCN}) using the graphs constructed by the methods themselves and those constructed by the RTIC. All the methods benefitted from training with the GCN-stream, which implies that the GCN-stream is a sufficiently general method that can be attached to any composition module. Moreover, the result shows that a higher graph quality consistently results in a more considerable gain, which implies that the single-model score can be improved with higher quality graphs.

\begin{table}[h!]
    \caption{The performance boost evaluated on Fashion-IQ when trained \textit{w/} or \textit{w/o} the GCN-stream. \textit{w/ GCN, Self} and \textit{w/ GCN, RTIC} indicate that the graph of the GCN-stream is constructed using the compared method itself and using the RTIC, respectively.}
    \centering
    \begin{adjustbox}{width=0.48\textwidth}
    \begin{tabular}{cccc||c}
    \toprule
    Method          & -         & w/ GCN, Self   & w/ GCN, RTIC     & $\textit{max.}\Delta$         \\ \hline \hline
    TIRG            & 30.71     & 31.26          & \textbf{31.39}   & \textcolor{green}{+2.21\%}    \\
    MRN             & 28.83     & 28.92          & \textbf{29.28}   & \textcolor{green}{+1.56\%}    \\
    ComposeAE       & 19.61     & \textbf{29.70} & 29.43            & \textcolor{green}{+33.97\%}   \\
    Param Hashing   & 26.54     & 26.86          & \textbf{26.89}   & \textcolor{green}{+1.31\%}    \\
    \bottomrule
    \end{tabular}
    \end{adjustbox}
    \label{tab:graph_quality}
\end{table}

\begin{figure}[h]
    \begin{center}
	    \includegraphics[width=1.0\linewidth]{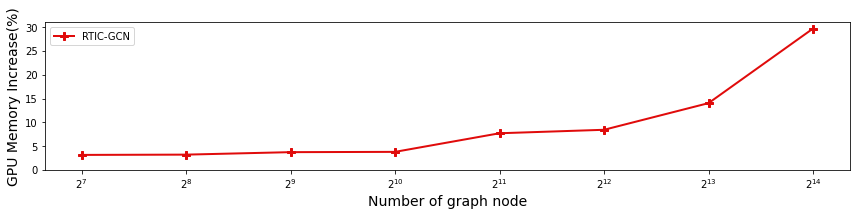}
	\end{center}
	\caption{Additional GPU memory required for GCN-stream at training phase. It is worth noting that the proposed GCN-stream requires no additional GPU memory at inference phase.}
	\label{fig:gpu_memory}
\end{figure}

\noindent
\textbf{Ablation study.}
This section describes the optimization process used for the challenge step by step and discusses how the trivial modification of training parameters can affect the final score.
The results are presented in Table~\ref{tab:ablation_study}. Based on the initial baseline score of 33.24, the single-model performance could be improved by up to 13\% by simply optimizing the training parameters. Following our optimization, we obtained a score of $37.18$ for TIRG, which outperforms the more recent methods. It strongly implies that trivial training details can significantly affect the overall performance. The result strongly highlights the importance of removing such undesired effects for a fair comparison.

    \begin{table}[h]
        \caption{The variation of the single-model performance with different training settings. The optimizer (from SGD to AdamW), the text encoder (from LSTM to 2-layered LSTM+GRU), and the embedding feature dimension of the composed feature (from 512 to 2048) are changed gradually to maximize the final score.}
        \centering
        \begin{adjustbox}{width=0.41\textwidth}
        \begin{tabular}{cc}
        \toprule
        Training details & Validation \\
        \hline \hline
        Baseline (RTIC) & 33.24 \\
        \hline
        + Replace optimizer & 34.89 (\textcolor{green}{+4.96\%}) \\
        + Increase embedding dimension & 36.79 (\textcolor{green}{+5.45\%}) \\
        + Replace text encoder & 37.72 (\textcolor{green}{+2.47\%}) \\
        + Spell correction & \textbf{38.22} (\textcolor{green}{+1.33\%}) \\
        \bottomrule
        \end{tabular}
        \end{adjustbox}
        \label{tab:ablation_study}
    \end{table}
    
\noindent
\textbf{Trade off.} Although training with the GCN-stream is a well-generalized technique that can consistently improve the score of any existing method as shown in Table~\ref{tab:graph_quality}, backpropagating gradients through the GCN layers requires additional GPU memory for training. Figure~\ref{fig:gpu_memory} shows the additional GPU memory required for joint training with the GCN-stream. We gradually increased the number of graph nodes and measured the relative increase in GPU memory used when the GCN-stream was attached to the main-stream. We report the relative rather than the absolute increases because the latter vary with the batch size and network complexity. In the inference phase, we no longer require the GCN layers for inference, which means that the GCN-stream does not slow down the inference speed at all.

\subsection{Visualization}

\begin{figure}[t]
    \begin{center}
	    \includegraphics[width=1.0\linewidth]{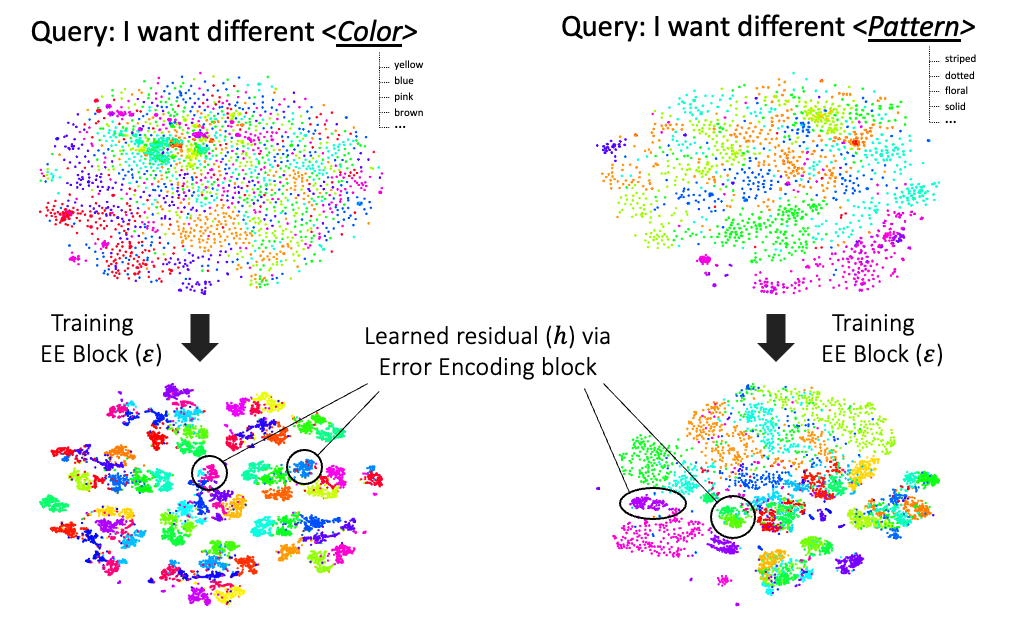}
	\end{center}
	\caption{t-SNE visualization of the feature distribution outputted by RTIC's error encoding block ($\mathcal{E}$).}
	\label{fig:tsne_error_encoding}
\end{figure}

Figure~\ref{fig:tsne_error_encoding} illustrates the t-SNE~\cite{van2008visualizing} visualization of the error-encoding block outputs from the RTIC. We sample 250 images from each of the eight different color classes (\textit{yellow, black, red, blue, green, brown, orange, pink}) and condition the images with the different class labels (\textit{e.g., blue image with the text ``yellow''}) instead of the captions for the query. We then extract the output of the error-encoding block and visualize the distribution. Note that images in the same class could have different attributes. For example, the images in the ``blue'' class will have the same color but different patterns (\textit{e.g., striped, dotted, or floral}). Nevertheless, Figure~\ref{fig:tsne_error_encoding} shows that the clusters are clearly formed, which indicates that the error encoding block successfully disentangles the other attributes from the representation when it is conditioned on the specific color-related text in the query. We also performed the same experiments with eight different pattern classes and observed consistent results.

%%%%%%%%% CONCLUSION
\section{Conclusion}
In this paper, we proposed a novel architecture, RTIC, specifically designed for the \textit{image-text composition} task. We also introduced a novel graph-based regularization technique attached to any composer in a plug-and-play manner. The joint training with the GCN-stream is a well-generalized approach in that it consistently improves the scores of existing composition methods. Finally, we achieved a state-of-the-art score on the \textit{image-text composition} task with the \textit{optimal environment} without exploiting any ensemble-like tricks such as aggregating features from the different stages or combining multiple loss functions. We also provide a fair and objective comparison between the composition methods by training models with \textit{unified} environment.

%%%%%%%%% REFERENCES
{\small
\bibliographystyle{ieee_fullname}
\bibliography{cvpr}
}

\newpage
\appendix

\section{Notation}
\begin{table}[H]
    \centering
    \begin{adjustbox}{width=0.45\textwidth}
    \begin{tabular}{cc}
        \toprule
        Symbol/Notation & Definition \\ 
        \hline \hline
        $I^{src}$ & Source image \\ 
        $I^{trg}$ & Target image \\
        $T$ & Text description \\
        $\psi(\cdot, \cdot)$ & Text-Image multi-modal composer \\
        $v^{src}$ & Visual feature extracted from source image \\
        $v^{trg}$ & Visual feature extracted from target image \\
        $\delta(\cdot, \cdot)$, $f(\cdot, \cdot)$ & Non-linear mapper \\
        $\Tilde{v}$ & Final composed feature \\
        $\mathcal{F}$ & Fusion block \\
        $\mathcal{E}$ & Error encoding block \\
        $\mathcal{G}$ & Gating block \\
        $n$ & Number of error encoding blocks in stack \\
        $\mathcal{X}^n$ & Output of the $n$-$th$ error encoding block in stack \\
        $K$ & Number of negative samples in mini-batch \\
        $B$ & Batch size \\
        $H$ & Node feature matrix \\
        $\Lambda$ & Correlation matrix \\
        $N$ & Number of nodes \\
        $D_V$ & Dimensionality of visual features \\
        $D_T$ & Dimensionality of text features \\
        $W$ & Transformation matrix \\
        $\mathcal{V}$ & Set of target image features \\
        $\lambda$ & Text encoder \\
        $\phi$ & Image encoder \\
        $\Lambda'$ & Binarized correlation matrix \\
        $\tau$ & Threshold for correlation matrix binarization \\
        $\Lambda''$ & Re-weighted correlation matrix \\
        $\hat{\Lambda}$ & Re-normalized correlation matrix \\
        $\Omega$ & Collection of image features \\
        $\mathcal{X}$ & Output of main stream \\
        $\mathcal{W}$ & Output of GCN stream \\
        $c$ & One hot vector \\
        \bottomrule
    \end{tabular}
    \end{adjustbox}
    \label{tab:notation}
\end{table}

\section{Architecture}

\begin{table}[H]
    \caption{Detailed configuration for RTIC architecture.}
    \centering
    \begin{adjustbox}{width=0.45\textwidth}
    \begin{tabular}{cccc}
        \toprule
        Module & Operation & Input dim.     & Output dim.    \\
        \hline \hline
        \multirow{4}{*}{Fusion block ($\mathcal{F}$)}         & Concat (dim=1) & $D_T$, $D_V$  & $D_T$ + $D_V$ \\
                                                              & BatchNorm1d    & $D_T$ + $D_V$ & $D_T$ + $D_V$ \\
                                                              & LeakyReLU      & $D_T$ + $D_V$ & $D_T$ + $D_V$ \\
                                                              & Linear         & $D_T$ + $D_V$ & $D_V$         \\
        \hline \hline
        \multirow{7}{*}{Error encoding block ($\mathcal{E}$)} & Linear         & $D_V$         & $D_V$ / 2     \\
                                                              & BatchNorm1d    & $D_V$ / 2     & $D_V$ / 2     \\
                                                              & LeakyReLU      & $D_V$ / 2     & $D_V$ / 2     \\
                                                              & Linear         & $D_V$ / 2     & $D_V$ / 2     \\
                                                              & BatchNorm1d    & $D_V$ / 2     & $D_V$ / 2     \\
                                                              & LeakyReLU      & $D_V$ / 2     & $D_V$ / 2     \\
                                                              & Linear         & $D_V$ / 2     & $D_V$         \\
        \hline \hline
        \multirow{5}{*}{Gating block ($\mathcal{G}$)}         & Linear         & $D_V$         & $D_V$         \\
                                                              & BatchNorm1d    & $D_V$         & $D_V$         \\
                                                              & LeakyReLU      & $D_V$         & $D_V$         \\
                                                              & Linear         & $D_V$         & $D_V$         \\
                                                              & Sigmoid        & $D_V$         & $D_V$         \\
        \bottomrule
    \end{tabular}
    \end{adjustbox}
    \label{tab:rtic_detail}
\end{table}

\section{Datasets}

Fashion-IQ~\cite{guo2019fashion} is a fashion product retrieval dataset that consists of images crawled from \textit{amazon.com} and natural language-based descriptions of one or more visual characteristics relating to the source and target images. It contains three categories, namely, dresses, toptees, and shirts. Following the same protocol as ~\cite{guo2019fashion}, we used 46,609 images and 18,000 triplets for training, and 5,373 images and 6,016 triplets for evaluation.

Shoes~\cite{guo2018dialog} is a dialog-based interactive retrieval dataset in which the images were originally crawled from \textit{like.com} and additionally tagged with natural language-based relative captions. Following~\cite{guo2018dialog}, we used 10,000 images and 8,990 triplets for training, and 4,658 images and 1,761 triplets for evaluation.

Birds-to-Words~\cite{forbes2019neural} consists of 3,520 bird images from \textit{iNaturalist} along with human-annotated natural language captions that describe the differences between pairs of images. Each text description is longer (average of 31 words) than the previous two datasets and has a more detailed description. We used 2,835 images and 12,805 triplets for training and 361 images and 1,556 triplets for evaluation.

\section{Learning Curves}
\begin{figure}[H]
    \begin{center}
	    \includegraphics[width=1.0\linewidth]{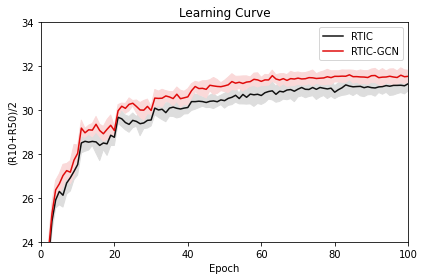}
	\end{center}
	\caption{Learning curves of RTIC and RTIC-GCN.}
	\label{fig:learning_curve}
\end{figure}

We compare the learning curves of RTIC when it is trained \textit{w/} or \textit{w/o} the GCN stream. The result is shown in Figure~\ref{fig:learning_curve}. The graph shows that the GCN stream can achieve a clear performance improvement against the other. Such tendency is consistent in that similar results are also observed using the other composition methods (\textit{e.g., TIRG}).

\section{Image and Text Representation}
Although the main focus of our work is on the composition of the two modalities, we observe that even a minor modification of the text and the image encoders dramatically affects the final performance. Therefore, we briefly describe our settings for learning the text and image representations.

For text representations, we learned sentence embedding from scratch using LSTM~\cite{hochreiter1997long}. Specifically, we tokenized the captions using a simple whitespace tokenizer and then built a word-to-vector vocabulary. Although applying GloVe~\cite{pennington2014glove} initialization on the word vectors can improve the result, we intentionally omitted the use of this technique and retained our training configuration as the baseline. The 1-layered LSTM forwarded the word embeddings, and then the outputs were max-pooled to obtain the sentence embedding. Finally, the outputs were projected onto the desired dimensions using a fully connected layer.

Global average pooling (GAP) was applied to the output of the last convolutional layer, followed by a fully connected layer. ResNet-50~\cite{he2016deep} was used as the backbone for the image encoder. We did not use a shallower backbone such as ResNet-18 because this made the performance difference between the composition methods relatively trivial, mainly due to the limited network capacity.

\section{Study on Other Possible Variants}
Our proposed GCN stream is a semi-supervised technique based on the structured graph. Since there exist various ways to utilize pseudo labels for training, we explore if other semi-supervised learning techniques can be employed for this task. We examine two different baselines to compare with GCN stream: \textit{Linear+BCE} and \textit{Pseudo Pairs}.

\noindent
\textbf{Linear+BCE.} Since the goal of the GCN stream is to learn a projection matrix $\mathcal{W} \in R^{D_V \times N}$, it is equivalent to learning a single-layer perceptron with no bias. Therefore, one can attach a linear layer instead of GCN layers behind the main-stream and use binary cross-entropy (BCE) loss with pseudo labels sampled from $A'$ as the GCN stream.

\noindent
\textbf{Pseudo Pairs.} Using the graph, we can also generate pseudo $\{I^{src}, T, I^{trg}\}$ pairs. The idea is that if two pairs have visually similar target image which has high cosine similarity above a certain threshold, we switch the $I^{trg}$ in two pairs. In this way, we double the number of $\{I^{src}, T, I^{trg}\}$ pairs in the train split of Fashion-IQ and train RTIC in the main-stream without GCN stream.

\begin{figure*}[t]
    \begin{center}
	    \includegraphics[width=1.0\linewidth]{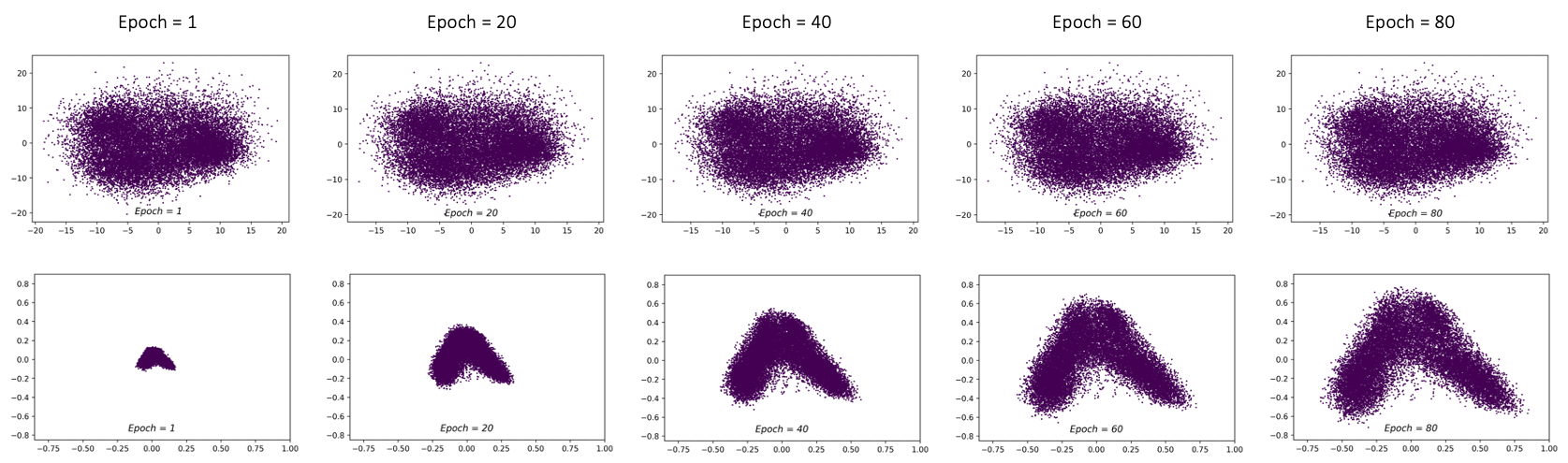}
	\end{center}
	\caption{PCA visualization of the feature distribution in the GCN stream for each epoch. The first and second rows represent the feature distribution in the GCN stream before and after it is forwarded by GCN layers respectively.}
	\label{fig:before_after_gcn}
\end{figure*}

The result is shown in Table~\ref{tab:semi_baseline}. The result shows that both \textit{Linear+BCE} and \textit{Pseudo Pairs} do not help training compared to the baseline that does not use any semi-supervised techniques. Only the GCN stream improves the result by 0.4\%p. It implies that simply applying BCE loss as auxiliary supervision does not improve the score (\textit{Linear+BCE}). Moreover, automatically generated pseudo pairs should be carefully considered because the noise degrades the score significantly (\textit{Pseudo Pairs}). This experiment shows that the performance gain due to the GCN stream originated from the use of binary classification loss for auxiliary supervision and a carefully designed GCN module.

    % Table 3. Other Possible Variants for semi-supervised learning
    \begin{table}[H]
    \caption{Comparison with other semi-supervised learning techniques. We measure (R@10 + R@50) / 2 for the score.}
    \centering
    \begin{adjustbox}{width=0.26\textwidth}
    \begin{tabular}{cccc}
        \toprule
        Method          & Score \\ \hline \hline
        Baseline        & 31.28 $\pm$ 0.22 \\
        Linear+BCE      & 27.44 $\pm$ 0.40 \\
        Pseudo Pairs    & 23.73 $\pm$ 0.18 \\
        GCN stream      & \textbf{31.68} $\pm$ \textbf{0.30} \\
        \bottomrule
    \end{tabular}
    \end{adjustbox}
    \label{tab:semi_baseline}
    \end{table}

\section{Analysis on Feature Distribution}
\noindent
\textbf{No weight sharing between composers.} The two composers in main-stream and GCN stream do not share the weights during the training. It is because the input for the GCN stream, $\mathcal{H}$, is static values while the input for the main-stream, which is the representations learned by the image and text encoders, is updated dynamically as the training proceeds. Such discrepancy between the input distributions of two composers degrades the performance if the weights are shared. Although we train two composers independently, the GPU memory required for training trivially increases because the composer has a small number of parameters.

\noindent
\textbf{Distribution before and after GCN layers.} Figure~\ref{fig:before_after_gcn} shows the PCA analysis on the feature distribution in the GCN stream before and after it is forwarded by GCN layers. The GCN layer converts the original feature distribution (\textit{first row}) extracted by the image and text encoders trained in the first stage into the more structured form of distribution (\textit{second row}), which implies the features are projected to be more distinguishable from each other. As the training proceeds, the features after the GCN layers are spread into space in a broader range.

\section{Naive comparison with feature fusion methods.}
We apply RTIC to the \textit{feature fusion} task in this section. The Table~\ref{tab:feature_fusion} shows the result on widely used benchmarks, VQA 2.0~\cite{goyal2017making} and VRD~\cite{lu2016visual}. We followed the training and testing environment with default parameters provided by the publicly released code~\footnote{https://github.com/Cadene/block.bootstrap.pytorch}. The result shows that the methods designed for the \textit{image text composition} task (\textit{TIRG} and \textit{RTIC}) are less effective for the \textit{feature fusion} task (\textit{BLOCK vs. RTIC}). Our interpretation is that since the \textit{image text composition} methods are forced to embed the composed feature into the image feature space, it is considered as a redundant constraint in the \textit{feature fusion} task, which degrades the performance.

% Table 4. feature fusion
    \begin{table*}[h!]
    \caption{PP: Phrase Prediction, PD: Phrase Detection, RD: Relationship Detection task. The comparison with the feature fusion methods on two benchmarks (VQA 2.0~\cite{goyal2017making} and VRD~\cite{lu2016visual}). We measure the accuracy for VQA 2.0 and R@50 for VRD.}
    \centering
    \begin{adjustbox}{width=0.88\textwidth}
    \begin{tabular}{ccccccccc}
    \toprule
     & \multicolumn{3}{c}{VRD} &  & \multicolumn{4}{c}{VQA 2.0 (\textit{test-std)}} \\ \cline{2-4} \cline{6-9} & Phrase prediction & Phrase detection & Relationship detection &  & All    & Yes/no   & Num   & Other \\ \hline \hline
    BLOCK & \textbf{86.46} & 22.96 & \textbf{19.26} & & \textbf{66.77} & \textbf{83.74} & \textbf{46.70} & \textbf{56.83} \\
    Mutan & 85.00 & 12.10 &  7.96 & &   -   &   -   &   -   &   -   \\
    MLB   & 84.85 & 19.08 & 15.55 & & 40.59 & 61.19 & 38.48 & 23.14 \\
    MFB   & 82.46 & 23.99 & 16.84 & & 38.50 & 61.18 & 32.35 & 20.16 \\
    MFH   & 83.42 & \textbf{24.01} & 16.89 & &   -   &   -   &   -   &   -   \\
    MCB   & 84.24 & 18.54 & 12.65 & & 41.65 & 73.83 & 34.66 & 15.24 \\ \hline
    TIRG  & 70.42 & 13.50 &  9.43 & & 54.03 & 73.75 & 34.24 & 41.39 \\
    RTIC  & 80.17 & 16.48 & 11.21 & & 47.00 & 61.17 & 26.01 & 39.49 \\
    \bottomrule
    \end{tabular}
    \end{adjustbox}
    \label{tab:feature_fusion}
    \end{table*}
    
    \begin{table*}[t]
    \caption{The detailed configuration of unified environment}
    \centering
    \begin{adjustbox}{width=1.0\textwidth}
    \begin{tabular}{cc}
        \toprule
        \multicolumn{2}{c}{Configuration} \\
        \hline \hline
        Image Encoder & ResNet50 (pretrained on ImageNet, output shape: 32 x 1024, avg-pooled) \\
        Text Encoder & LSTM (num\_layers: 1, input\_size: 1024, hidden\_size: 1024, output shape: 32 x 1024, max-pooled) \\
        Image Size / Crop Size & 256 / 224 \\
        Image Transform (Train) & ResizeWithPadding, RandomCrop, RandomHorizontalFlip, RandomAffine, Normalize\\
        Image Transform (Test) & ResizeWithPadding, CenterCrop, Normalize \\
        Optimizer & SGD \\
        Loss & DML loss, K = B~\cite{vo2019composing} \\
        Training Epochs & 100 \\
        Learning rate & 0.01 \\
        Learning rate decay & policy: StepLR, factor: $1/\sqrt{2}$, step\_size: 10 \\
        Others & No spell correction, No initialization with GloVe  \\
        Metric & Average Recall@K over 8 trials  \\
        \bottomrule
    \end{tabular}
    \end{adjustbox}
    \label{tab:training_standard}
\end{table*}
    
\section{Score Inconsistency across Papers}
Table~\ref{tab:inconsistency} shows the inconsistency in reported scores across papers. We mainly investigate the score of TIRG~\cite{vo2019composing} because it is the most widely used method for comparison in various studies. We found a significant gap between the reported scores, which is the reason we address the need for a \textit{unified environment} for training to ensure objectiveness and fairness in performing comparisons across papers.

\section{Unified Training Environment}
In Table~\ref{tab:training_standard}, we suggest a baseline for training to accomplish a fair comparison. We keep the image and text encoder as simple as possible while keeping the complexity of encoders high enough to represent the visual and contextual information. The image is resized to $256 \times 256$ with padding while keeping the ratio, then cropped to $224 \times 224$. We use DML loss (K=B) suggested in \cite{vo2019composing}, mainly because it can suppress undesirable effects caused by the sampling strategy. According to the technical reports of the retrieval challenge, spell correction increases the result significantly.
Moreover, initializing word embeddings with GloVe or BERT provides better generalization when the embeddings are fine-tuned. However, we omit such auxiliary techniques to keep our training environment simple. We strongly suggest using the metrics averaged over eight trials because the scores severely fluctuate even between models trained with the same training parameters.

\begin{table}[H]
    \caption{TIRG scores reported across papers.}
    \centering
    \begin{adjustbox}{width=0.45\textwidth}
    \begin{tabular}{cccc}
    \toprule
    (R@10+R@50)/2 & Fashion-IQ & Fashion200K & Shoes \\ \hline \hline
    ~\cite{chen2020image}           & 27.39 & 53.90 & 57.42 \\
    ~\cite{anwaar2020compositional} & 20.15 & 53.15 & - \\
    ~\cite{jandial2020trace}        & 19.95 & - & 38.79 \\
    ~\cite{chen2020learning}        & 15.27 & 51.95 & - \\ \hline
    $\textit{max.}\Delta$           & \textbf{12.13} & \textbf{1.95} & \textbf{18.63} \\
    \bottomrule
    \end{tabular}
    \end{adjustbox}
    \label{tab:inconsistency}
\end{table}

\section{Qualitative Result}
\begin{figure*}[t]
    \begin{center}
	    \includegraphics[width=0.84\linewidth]{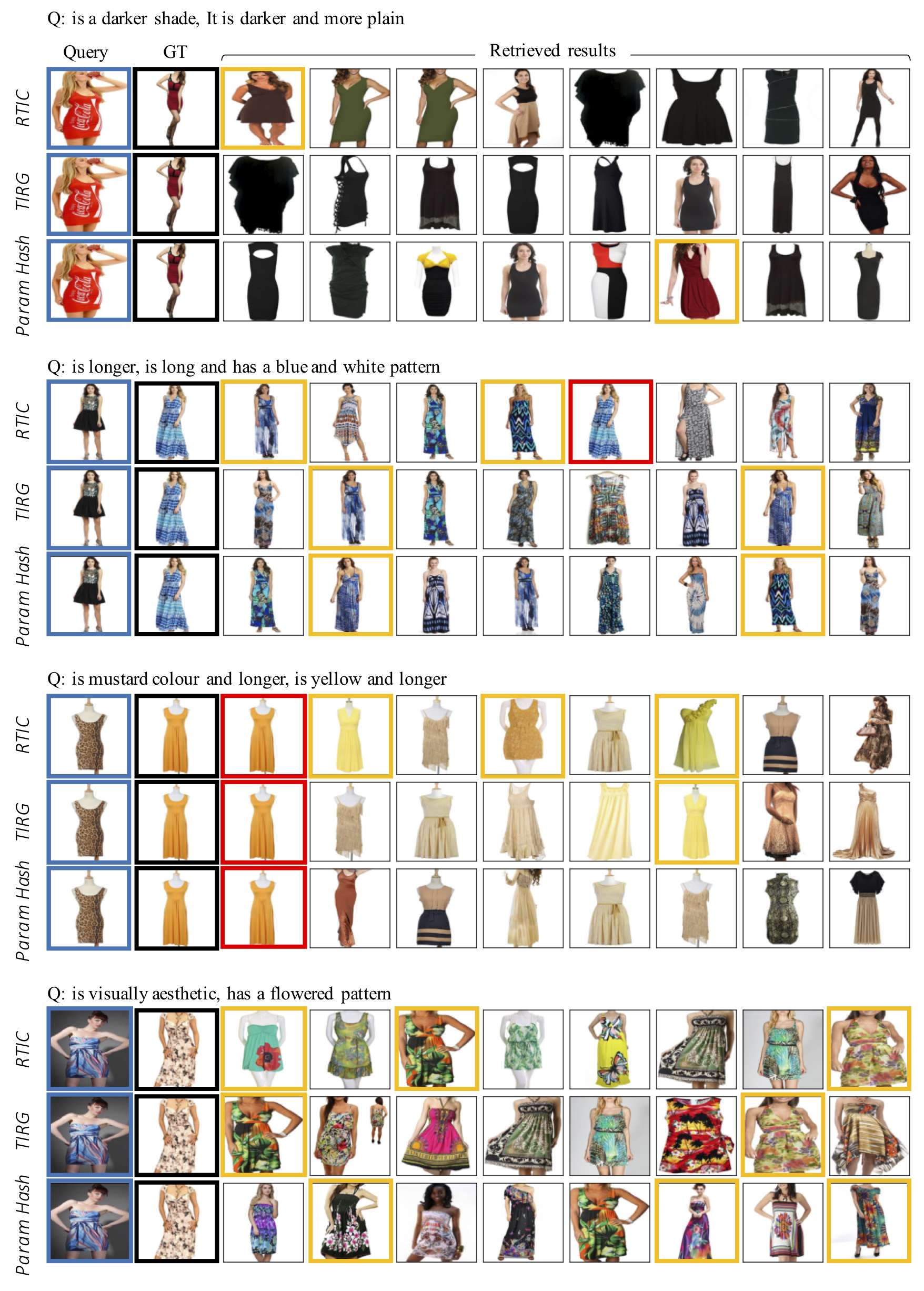}
	\end{center}
	\caption{Top-8 retrieval examples of RITC, TIRG, and Parameter Hashing. The blue and the black border indicate the query and the ground-truth respectively. The red border means the answer is correctly found. The orange border is means the image is not the answer but perceptually acceptable. }
	\label{fig:retrieval}
\end{figure*}

The image retrieval result of the top 8 candidates is illustrated in Figure~\ref{fig:retrieval}. The examples show that even though a ground-truth $I^{src} T, I^{trg}$ pair is defined, the text can not fully describe the fundamental aspects of the target image. Therefore, the answer can be multiple similarly-looking items that are perceptually acceptable. Ideally, the model is guided to obtain clues for the other attributes that are not described by the text based on the source image.

\end{document}